\documentclass[twoside]{article}

\usepackage{fitee}

\usepackage[colorlinks, breaklinks = true]{hyperref}		
\usepackage[utf8]{inputenc} 
\usepackage[T1]{fontenc}    
\usepackage{url}            
\usepackage{booktabs}       
\usepackage{amsfonts}       
\usepackage{nicefrac}       
\usepackage{microtype}      
\usepackage{xcolor}         
\usepackage{amsmath} 
\usepackage{tcolorbox}
\usepackage{enumitem}
\usepackage{multirow}
\usepackage{array}
\usepackage{multirow}
\usepackage{makecell}
\usepackage{array}
\usepackage{graphicx}
\usepackage{tabularx}
\usepackage{subfigure}

\usepackage[table]{xcolor}
\definecolor{topone}{RGB}{255,235,235}
\definecolor{toptwo}{RGB}{235,245,255}

\newcommand{\best}[1]{\cellcolor{topone}\textbf{#1}}
\newcommand{\second}[1]{\cellcolor{toptwo}\underline{#1}}

\addto{\captionsenglish}{%
  
}

\begin{document}

\title{PoliLegalLM: A Technical Report on a Large Language Model for Political and Legal Affairs}

\author[$\dagger$1]{Yu-ting HUANG}%
\author[$\dagger$1]{Ying-hao HU}%
\author[2]{Qian XIAO}%
\author[1]{Wen-lin ZHONG}%
\author[$\ddagger$1]{Yi-quan WU}%
\author[3]{\\Tai-shi ZHOU}%
\author[3]{Mo-ke CHEN}%
\author[1]{Chang-long SUN}%
\author[$\ddagger$1]{Kun KUANG}%
\author[1]{Fei WU}

\affil[1]{Zhejiang University, Hangzhou, China}
\affil[2]{Tongyi Lab, Alibaba Group, Hangzhou, China}
\affil[3]{Zhejiang Zhengfa Xinxi Guanli Zhongxin, Hangzhou, China}



\authmark{}



\corremailA{wuyiquan@zju.edu.cn, kunkuang@zju.edu.cn}
\emailmark{$\ddagger$}	

\dateinfo{}

\abstract{Large language models (LLMs) have achieved remarkable success in general-domain tasks, yet their direct application to the legal domain remains challenging due to hallucinated legal citations, incomplete knowledge coverage, and weak structured reasoning. To address these issues, we propose \textbf{PoliLegalLM}, a domain-specific large language model tailored for political and legal applications. Our approach adopts a unified training framework that integrates continued pretraining, progressive supervised fine-tuning, and preference-based reinforcement learning to jointly enhance legal knowledge grounding, task alignment, and reasoning capability. We construct a large-scale, high-quality legal corpus and design a structured post-training pipeline, enabling the model to effectively learn domain-specific knowledge and adapt to diverse legal tasks. We evaluate PoliLegalLM on three representative benchmarks, including LawBench, LexEval, and a real-world dataset, PoliLegal. Experimental results demonstrate that PoliLegalLM achieves strong and consistent performance, outperforming competitive models of similar scale and remaining highly competitive with significantly larger models, while achieving the best results on real-world legal scenarios. These results highlight the effectiveness of our training paradigm and the practical value of domain-specific LLMs for real-world legal applications.}






\articleType{}

\maketitle

\section{Introduction}

Recent advances in large language models (LLMs) have significantly improved performance across a wide range of natural language processing tasks~\cite{zhao2026surveylargelanguagemodels, brown2020languagemodelsfewshotlearners,openai2024gpt4technicalreport,zhang2025surveyreinforcementlearninglarge,zhang2026landscapeagenticreinforcementlearning}. However, directly applying general-purpose LLMs to the legal domain remains highly challenging. Legal reasoning requires not only extensive domain knowledge but also strict logical consistency, precise statutory grounding, and reliable decision-making processes~\cite{dahl2024large,guha2023legalbenchcollaborativelybuiltbenchmark,cui2024chatlawmultiagentcollaborativelegal}. In practice, existing models often suffer from hallucinated legal citations, incomplete or outdated knowledge coverage, and weak structured reasoning, which severely limits their applicability in judicial and governance scenarios~\cite{Hu_2025,zhang2025explicitsyllogisticlegalreasoning,kalra2025hyparaghybridparameteradaptive}.

While post-training foundation models on legal-domain data have shown promising improvements~\cite{chi2026legalai, colombo2024saullm7bpioneeringlargelanguage,yue2023disclawllmfinetuninglargelanguage}, such approaches remain insufficient for real-world legal applications. In particular, existing methods often rely on data with limited coverage and standard fine-tuning paradigms, which fail to adequately capture statutory knowledge and its complex interdependencies. Consequently, these models struggle to achieve both comprehensive knowledge coverage and consistently high-quality outputs. Therefore, adapting LLMs to the legal domain requires a unified approach that jointly addresses two fundamental challenges: 
(1) scaling and structuring domain-specific data to capture complex legal knowledge and its interdependencies; and 
(2) designing training strategies that promote accurate, consistent, and reliable outputs.

In this work, we propose a systematic approach for constructing high-quality training data, and introduce \textbf{PoliLegalLM}, a domain-specific large language model designed for political and legal applications. 

First, we construct a large-scale, high-quality legal corpus that integrates diverse data sources, including judicial judgments, statutory texts, legal literature, and real-world governance data. The continued pretraining corpus comprises 140B tokens, while the post-training stage consists of 1,834,198 instruction samples. To ensure data reliability and domain suitability, we adopt a multi-stage construction strategy, including iterative filtering and LLM-based quality scoring. Furthermore, we augment key legal texts with structured reasoning annotations, which enhance knowledge representation and improve downstream task performance.

Building on this corpus, we design a unified three-stage training framework consisting of continued pretraining (CPT), Progressive Supervised Fine-Tuning (PSFT), and Hard Sample-aware Iterative Direct Preference Optimization (HIPO). Specifically, PSFT follows a curriculum-style paradigm: the model is first trained on a core task, legal judgment prediction, to learn structured mappings from case facts to legal conclusions, and is then progressively adapted to diverse downstream tasks. To ensure that core legal knowledge and reasoning skills are preserved during this progression, PSFT explicitly incorporates a lightweight mixed-task training mechanism as an anti-forgetting strategy, enabling the model to retain previously acquired capabilities while adapting to new tasks. On top of this, HIPO focuses on hard samples where the model underperforms and applies preference learning to iteratively refine model behavior, thereby improving factual consistency and output reliability. Meanwhile, an auxiliary negative log-likelihood (NLL) objective is integrated into the preference optimization process, providing a stabilizing signal that reinforces high-quality responses and mitigates potential degradation during iterative updates.

To comprehensively evaluate our approach, we benchmark PoliLegalLM on three representative datasets: \textbf{LawBench}, \textbf{LexEval}, and \textbf{PoliLegal}. These benchmarks cover a wide spectrum of legal capabilities, including knowledge memorization, reasoning, generation, and real-world task performance. Experimental results demonstrate that PoliLegalLM achieves strong performance across multiple benchmarks, outperforming competitive small- and medium-scale models and remaining highly competitive with significantly larger models, while achieving the best results on the real-world PoliLegal benchmark. These gains are driven by effective domain-specific training, which consistently improves both legal knowledge coverage and output quality across knowledge-oriented and reasoning-intensive tasks. Ablation studies further validate the effectiveness of our three-stage training pipeline.
\section{Related Work}

This section reviews recent advances in LLMs for the legal domain, with a focus on methods for adapting models to legal tasks and improving the quality and reliability of their outputs.

\subsection{Instruction-Following Legal LLMs}

In recent years, LLMs for the legal domain have generally followed a two-stage paradigm, consisting of domain-adaptive pre-training and post-training. Representative legal LLMs, including LawGPT~\cite{zhou2024lawgptchineselegalknowledgeenhanced}, HanFei~\cite{HanFei}, and InternLM-Law~\cite{fei2024internlmlawopensourcechinese}, primarily rely on CPT and SFT to adapt general-purpose models to legal scenarios, achieving notable improvements over base models on a variety of downstream legal tasks. Subsequent studies further explore instruction tuning~\cite{shu2024lawllm}, large-scale fine-tuning with legal syllogism–oriented data~\cite{yue2023disclawllmfinetuninglargelanguage}, or architectural extensions such as Mixture-of-Experts to enhance model capacity and data coverage~\cite{cui2024chatlawmultiagentcollaborativelegal}. Despite these advances, existing legal instruction models heavily depend on large-scale annotated datasets and predominantly focus on aligning models to surface-level task objectives. 

\subsection{Reasoning Legal LLMs}

Recently, many LLMs have begun to explicitly generate intermediate reasoning, planning, or problem decomposition traces prior to producing final answers, aiming to improve reliability and structural coherence. Representative reasoning-oriented models demonstrate that explicit reasoning can enhance performance on complex, non-knowledge-intensive legal tasks~\cite{Hu_2025}. In the legal domain, several models have explored explicit legal reasoning through reinforcement learning–based training strategies. LexPam~\cite{zhang2025legalmathematicalreasoningllms} applies a two-stage GRPO framework to legal numerical reasoning, but its scope remains limited to narrowly defined tasks. SyLeR~\cite{zhang2025explicitsyllogisticlegalreasoning} promotes explicit syllogistic reasoning, yet relies on similarity-based evaluation signals that fail to adequately penalize legally invalid but semantically plausible outputs. LexPro~\cite{chen2025lexpro10technicalreport} combines supervised fine-tuning with reinforcement learning to refine reasoning behaviors, but does not incorporate reward functions specifically grounded in legal constraints. Unilaw-R1~\cite{cai2025unilawr1largelanguagemodel} introduces correctness-oriented reward dimensions, which in knowledge-intensive legal settings may propagate the biases and blind spots of imperfect adjudication signals. 
\section{Corpus Construction}

To support training of PoliLegalLM, we construct a comprehensive and diversified data pool that integrates both general-domain and legal-domain resources.

\subsection{CPT Corpus Construction}

\subsubsection{CPT Data Sources}

Our training data is collected from diverse sources and categorized as follows:

\begin{itemize}[
  leftmargin=1.1em,
  labelsep=0.4em,
  itemsep=0.3em,
  topsep=0pt,
  parsep=0pt,
  partopsep=0pt
]
  \item \textbf{General Industry Corpora}: multilingual Chinese--English corpora covering major industrial domains, following the construction principles of IndustryCorpus2~\citep{beijing_academy_of_artificial_intelligence}, designed to provide broad linguistic coverage and cross-domain generalization ability.
  \item \textbf{Legal and Political News}: professionally written news articles covering legislation, judicial activities, public governance, and legal policy.
  \item \textbf{Judicial Judgment Documents}: court judgments from multiple jurisdictions and case types, including factual descriptions, judicial reasoning, and adjudicative outcomes.
  \item \textbf{Legal Books and Academic Papers}: textbooks, monographs, and scholarly publications that systematically present legal doctrines, theories, and research findings.
  \item \textbf{Articles and Judicial Interpretations}: legal texts, including statutes, judicial interpretations, local regulations, and departmental rules.
\end{itemize}

\subsubsection{CPT Data Processing}
\label{sec:CPTDataProcessing}

For the collected raw data, we adopt a three-stage data processing pipeline to ensure both data quality and computational efficiency. The pipeline consists of three key stages: \textbf{filtering, scoring, and enhancement}.

\textbf{Filtering.} We first apply rule-based data cleaning to remove noisy and low-quality samples from the corpus. Specifically, we filter out texts that are excessively short or overly long, as they often correspond to incomplete fragments or improperly concatenated documents. We also remove samples dominated by special characters, markup symbols, or formatting artifacts, which typically arise from web scraping or parsing errors and do not contain meaningful linguistic information.

\begin{figure}[t!]
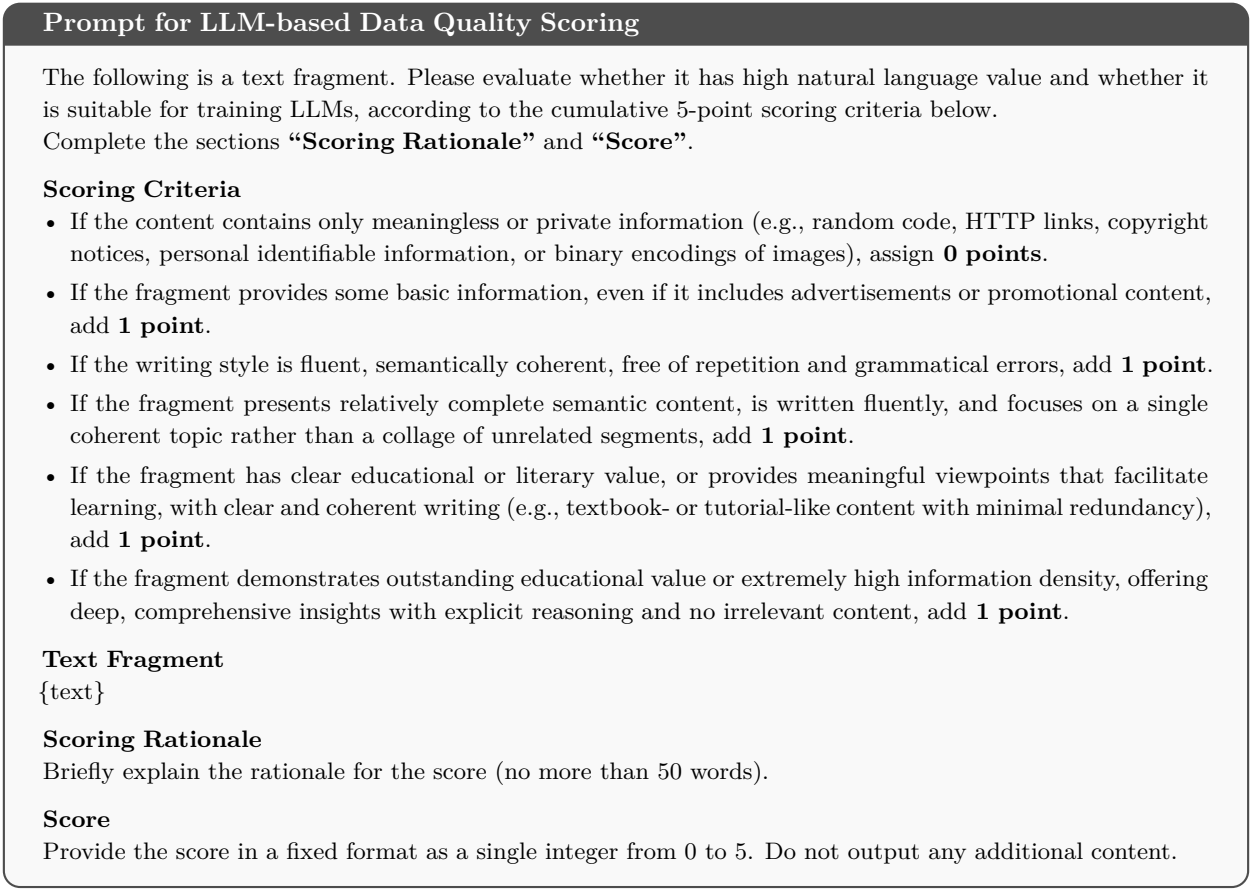

\centering
\begin{tcolorbox}[
  colback=gray!5,
  colframe=black!70,
  title=\textbf{Prompt for LLM-based Data Quality Scoring},
  fonttitle=\bfseries,
  boxrule=0.8pt,
  arc=2mm
]
\small
The following is a text fragment. Please evaluate whether it has high natural language value and whether it is suitable for training LLMs, according to the cumulative 5-point scoring criteria below.

Complete the sections \textbf{``Scoring Rationale''} and \textbf{``Score''}.

\medskip
\textbf{Scoring Criteria}

\begin{itemize}[
  leftmargin=1.1em,
  labelsep=0.4em,
  itemsep=0.3em,
  topsep=0pt,
  parsep=0pt,
  partopsep=0pt
]
  \item If the content contains only meaningless or private information (e.g., random code, HTTP links, copyright notices, personal identifiable information, or binary encodings of images), assign \textbf{0 points}.
  \item If the fragment provides some basic information, even if it includes advertisements or promotional content, add \textbf{1 point}.
  \item If the writing style is fluent, semantically coherent, free of repetition and grammatical errors, add \textbf{1 point}.
  \item If the fragment presents relatively complete semantic content, is written fluently, and focuses on a single coherent topic rather than a collage of unrelated segments, add \textbf{1 point}.
  \item If the fragment has clear educational or literary value, or provides meaningful viewpoints that facilitate learning, with clear and coherent writing (e.g., textbook- or tutorial-like content with minimal redundancy), add \textbf{1 point}.
  \item If the fragment demonstrates outstanding educational value or extremely high information density, offering deep, comprehensive insights with explicit reasoning and no irrelevant content, add \textbf{1 point}.
\end{itemize}

\medskip
\textbf{Text Fragment}

\{text\}

\medskip
\textbf{Scoring Rationale}

Briefly explain the rationale for the score (no more than 50 words).

\medskip
\textbf{Score}

Provide the score in a fixed format as a single integer from 0 to 5. Do not output any additional content.
\end{tcolorbox}
\caption{Prompt used for LLM-based data quality scoring during corpus construction.}
\label{fig:corpus_construction_llm}
\end{figure}

\textbf{Scoring.} To enable scalable quality assessment beyond rule-based filtering, we construct a lightweight scoring model to estimate the quality of each text fragment. Specifically, we first randomly sample 10{,}000 instances from the candidate corpus and annotate them using a strong LLM (see Fig.~\ref{fig:corpus_construction_llm}). The scoring rubric ranges from 0 to 5, where higher scores indicate better semantic completeness, linguistic fluency, topical coherence, and educational or informational value. Based on these LLM-annotated samples, we then train a lightweight scoring model, initialized from Qwen3-0.6B~\citep{yang2025qwen3technicalreport}, to predict quality scores for the remaining corpus at scale. The resulting scorer demonstrates strong agreement with human judgments, achieving a Spearman’s $\rho$ of 0.97, a mean absolute error (MAE) of 0.42, and an adjacent accuracy (within $\pm 1$) of 0.95.

\textbf{Enhancement.} Finally, for high-value legal texts (e.g., statutes), we adopt a semi-automatic data generation pipeline. Specifically, we collaborate with legal practitioners to formalize a multi-dimensional knowledge framework for legal statutes, covering key aspects such as normative rules, legal elements and consequences, conceptual definitions, systemic relations, and applicability boundaries. These dimensions reflect the inherent semantic structure of statutory provisions and serve as the foundation for controllable data generation. Based on this schema, we design a unified instruction template that explicitly encodes each knowledge dimension as a generation constraint (see Appendix~\ref{appendix:prompt_template} for details). Given a legal provision, the template prompts a strong LLM to perform dimension-wise analysis and synthesize high-quality instruction--response pairs conditioned on predefined knowledge dimensions, thereby encouraging structured reasoning chains and legally grounded explanations. Representative examples are provided in Table~\ref{tab:instruction_example}.

\subsubsection{CPT Corpus Statistics}

Using the above pipeline, we construct a CPT corpus of \textbf{140B tokens} from an initial pool of 445B raw tokens. All corpora used in the CPT stage are uniformly converted into continuous text sequences in a unified \textbf{\{text\}} format. Meanwhile, we carefully balance language coverage and domain specificity. Specifically, the corpus maintains a Chinese--English ratio of 7:3 and a domain--general ratio of 6:4, ensuring both strong domain grounding and general linguistic robustness. Table~\ref{tab:cpt_data} summarizes the data composition used for CPT, including corpus type, number of documents, total tokens, and sampled tokens.

\begin{table}[h]
\centering
\small
\caption{Composition of the corpus used for CPT.}
\vspace{4pt}
\begin{tabular}{>{\centering\arraybackslash}m{1.8cm} l r r r}
\toprule
\textbf{Language} & \textbf{Corpus Type} & \textbf{\# Documents} & \textbf{Tokens} & \textbf{Sampled Tokens} \\
\midrule
\multirow{2}{*}{English}
 & General Industry Corpora & 57.9M & 138B & 20B \\
 & Legal and Political News   & 22.5M & 39B  & 20B \\
\midrule
\multirow{5}{*}{Chinese}
 & General  Industry Corpora & 24.3M & 87B  & 35B \\
 & Legal and Political News & 9.6M  & 22B  & 11B \\
 & Judicial Judgment Documents  & 86.4M & 155B & 50B \\
 & Articles and Judicial Interpretations & 1.55M & 2B & 2B \\
 & Legal Books and Academic Papers  & 0.13M & 2B   & 2B  \\
\midrule
\textbf{Total} & & 202.4M & 445B & \textbf{140B} \\
\bottomrule
\end{tabular}
\label{tab:cpt_data}
\end{table}

\subsection{Post-training Corpus Construction}

\subsubsection{Post-training Data Sources}

The post-training corpus is designed to complement the pretraining data by emphasizing high-quality, task-oriented samples, with a particular focus on legal and governance-related scenarios.

We organize the construction of our post-training data into three hierarchical levels:

\begin{itemize}[
  leftmargin=1.1em,
  labelsep=0.4em,
  itemsep=0.3em,
  topsep=0pt,
  parsep=0pt,
  partopsep=0pt
]

\item \textbf{Level 01: General Dialogue Corpus.} 
This level establishes the model’s foundational conversational capabilities, enabling it to acquire broad knowledge and respond fluently to diverse open-ended user queries. The data is primarily sourced from \textit{Infinity-Instruct}~\cite{li2025infinityinstructscalinginstruction} and \textit{Chinese Fineweb Edu}~\cite{yu2025opencsgchinesecorpusseries}.

\item \textbf{Level 02: General Instruction-Following Corpus.} 
This level focuses on enhancing the model’s ability to follow structured instructions and perform task-oriented reasoning. It includes a wide range of general instruction data, such as information extraction, summarization, classification, and multi-step problem solving, bridging the gap between open-ended dialogue and specialized domain tasks. The data is mainly collected from \textit{MSAgent-Bench}~\cite{li2023modelscopeagentbuildingcustomizableagent} and \textit{Chinese-Qwen3-235B-2507-Distill-data-110k-SFT}~\cite{ qwen2025_distill}.

\item \textbf{Level 03: Political and Legal Corpus.} 
This level focuses on domain-specific expertise and consists of three key components: 
\textit{(i) Article Memory}, which involves learning national and Zhejiang provincial legal provisions for precise legal retrieval and citation; 
\textit{(ii) Political and Legal Tasks}, which decompose real-world business scenarios to train capabilities such as document proofreading and case analysis; and 
\textit{(iii) Document Generation}, which leverages standardized templates and business logic to generate compliant legal documents. 
The data is derived from real-world sources, including anonymized internal datasets and domain-specific proprietary resources, such as service logs, police incident reports, and judicial case documents.
\end{itemize}

\subsubsection{Post-training Data Processing}

We adopt a pipeline to ensure the quality, consistency, and balance of the post-training corpus.

\textbf{Political and Legal Corpus Processing.}
For domain-specific data, we first collaborate with legal experts to pre-annotate a subset of samples and define standardized formats for different sub-tasks. Based on this, we apply rule-based processing to convert raw data into structured \{query, golden answer\} pairs. We further employ the scoring model described in Section~\ref{sec:CPTDataProcessing} to assess data quality, filtering out samples below a predefined threshold.

\textbf{Data Proportion Control.}
We maintain an approximate 7:3 ratio between general-purpose and domain-specific data, balancing general instruction-following capability with domain specialization.

\subsubsection{Post-training Corpus Statistics}

Our post-training data consists of SFT data and reinforcement learning (RL) preference data. In the SFT stage, samples are organized as \textbf{\{query, golden answer\}} pairs. For the RL stage, we construct preference pairs by reusing hard samples from the SFT data, formatted as \textbf{\{query, chosen answer, rejected answer\}}. Specifically, the golden answer is treated as the chosen response, while low-quality model-generated outputs are used as rejected responses. More details can be found in Section~\ref{sec:ReinforcementLearning}. In total, the post-training corpus contains \textbf{1,834,198} instruction samples. As shown in Table~\ref{tab:post_training_data}, it includes \textbf{224,465} English general-domain samples (12.5\%), \textbf{991,681} Chinese general-domain samples (55.6\%), and \textbf{618,052} Chinese political and legal domain samples (31.7\%).

\begin{table}[h]
\centering
\small
\caption{Composition of the post-training instruction fine-tuning corpus.}
\vspace{4pt}
\begin{tabular}{>{\centering\arraybackslash}m{4cm} m{4.5cm} c}
\toprule
\textbf{Category} & \textbf{Corpus Type} & \textbf{\# Samples} \\
\midrule
\multirow{2}{*}{\shortstack{English \\ General Corpus}}
 & Dialogue QA  & 198,847 \\
 & Instruction Following & 25,618 \\
\cmidrule{1-3}
\multirow{2}{*}{\shortstack{Chinese \\ General Corpus}}
 & Dialogue QA & 589,226 \\
 & Instruction Following  & 402,455 \\
\cmidrule{1-3}
\multirow{4}{*}{\shortstack{Chinese \\ Political \& Legal Corpus}}
 & Dialogue QA  & 83,929 \\
 & Article Memory  & 67,913 \\
 & Political \& Legal Tasks  & 383,953 \\
 & Document Generation  & 82,257 \\
\midrule
\textbf{Total} & & \textbf{1,834,198} \\
\bottomrule
\end{tabular}
\label{tab:post_training_data}
\end{table}
\section{Model Training}

\begin{figure}[t!]
    \centering
    \includegraphics[width=\linewidth]{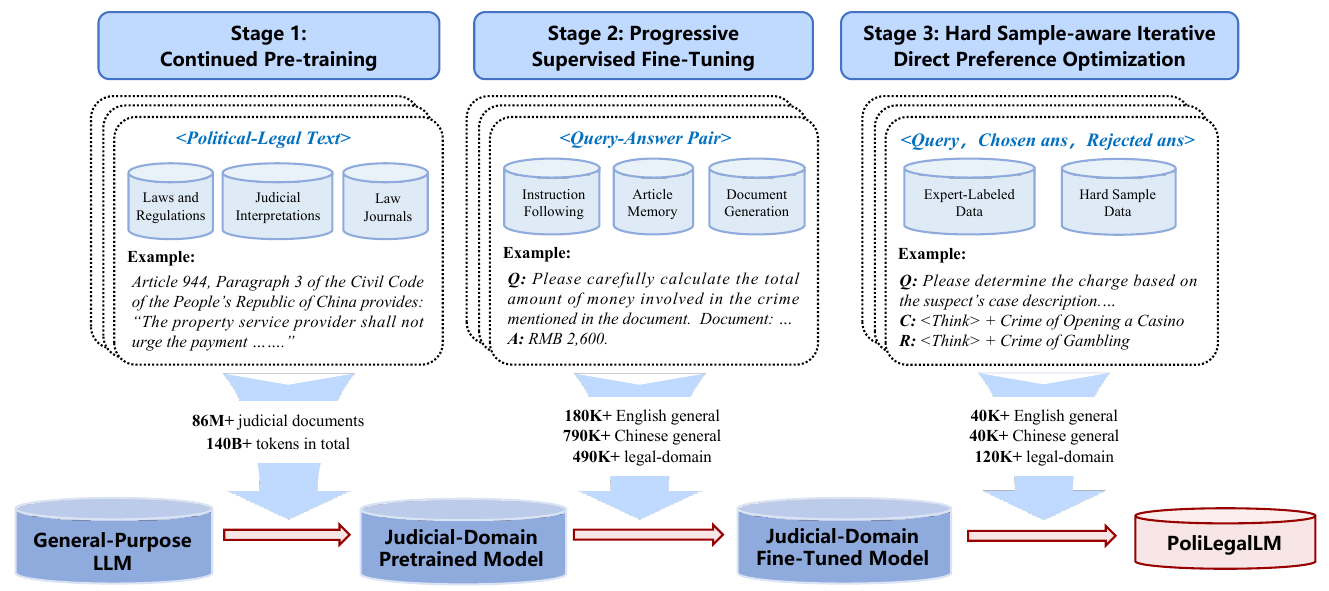}
    \caption{Overview of the multi-stage training pipeline, including continued pretraining, supervised fine-tuning, and reinforcement learning.}
    \label{fig:workflow}
\end{figure}

To fully leverage the constructed corpus and enable effective domain adaptation, we adopt a multi-stage training pipeline consisting of continued pretraining, supervised fine-tuning, and reinforcement learning, as illustrated in Figure~\ref{fig:workflow}. These stages are designed to progressively enhance domain knowledge acquisition, task alignment, and reasoning capability, forming a unified training framework for legal language modeling.

\subsection{Continued Pretraining}
\label{sec:cpt}

CPT serves as the primary mechanism for injecting domain-specific knowledge into the base model. Starting from a strong general-purpose foundation, CPT allows the model to systematically absorb legal and political knowledge, including statutory language, judicial reasoning patterns, and domain-specific discourse structures. Compared with instruction tuning alone, CPT provides deeper and more stable integration of knowledge at the representation level, thereby improving downstream performance in high-stakes domain reasoning tasks. To balance training efficiency and long-context adaptation, we adopt a two-stage CPT strategy.

\begin{itemize}[
  leftmargin=1.1em,
  labelsep=0.4em,
  itemsep=0.3em,
  topsep=0pt,
  parsep=0pt,
  partopsep=0pt
]
  \item \textbf{Stage I} performs training on a 8K-token context window using 90\% of the CPT data. This stage focuses on efficiently injecting large-scale legal knowledge while maintaining manageable computational cost.
  
  \item \textbf{Stage II} switches to an 16K-token context window and uses the remaining 10\% of the data, enabling the model to adapt to long-context legal documents such as judgments and statutes. Across both stages, we ensures a constant throughput of approximately 786K tokens per optimization step.
\end{itemize}

At the beginning of Stage II, we apply a short warmup schedule with a learning rate that is approximately continuous with, but slightly higher than, the terminal learning rate of Stage I. This design stabilizes optimization while allowing rapid adaptation to the longer context length.

\subsection{Supervised Fine-Tuning}

After CPT on legal and judicial corpora, the model acquires extensive domain-specific knowledge, including legal terminology, statutory provisions, and case-related patterns. However, such knowledge is primarily learned in a \emph{passive} manner and is not directly aligned with the structured reasoning and decision-making processes required in real-world judicial scenarios. As a result, models trained solely with CPT often struggle to effectively apply legal knowledge to practical tasks, especially those requiring structured prediction and multi-step reasoning.

To address this limitation, we introduce a SFT stage to explicitly align the model with practical legal tasks. Within this stage, we observe that directly fine-tuning on heterogeneous downstream tasks may lead to unstable optimization and suboptimal generalization. To mitigate this issue, we propose a \textbf{Progressive Supervised Fine-Tuning (PSFT)}~\cite{Hu_2025} strategy, which organizes SFT in a curriculum-style manner, gradually transferring foundational legal knowledge to diverse downstream applications. Furthermore, to prevent the degradation of core legal knowledge during progressive adaptation, we incorporate a lightweight \textbf{anti-forgetting mechanism} based on mixed-task training as part of PSFT. Together, these components form a unified SFT framework, enabling improved downstream performance over standard SFT, while mitigating catastrophic forgetting during progressive adaptation.

\paragraph{PSFT Strategy.}
PSFT follows a two-stage curriculum-style training paradigm, consisting of (1) core skill learning and (2) progressive adaptation. Specifically, we first fine-tune the model on a fundamental legal task, Legal Judgment Prediction, which includes charge prediction, applicable law identification, and penalty estimation. These sub-tasks correspond to the core decision-making pipeline in real judicial practice, establishing a structured mapping from case facts to legal conclusions.

Importantly, Legal Judgment Prediction serves as a prerequisite task for a wide range of downstream applications. Many legal tasks either directly rely on or are implicitly derived from these core predictions. For instance, legal question answering and consultation require identifying applicable laws and reasoning over potential charges; legal document generation depends on accurate charge and penalty determination; and similar case retrieval relies on consistent judgment outcomes as structured anchors. Therefore, learning this task equips the model with essential reasoning patterns and supervisory signals that are broadly reusable across downstream scenarios. This stage produces an intermediate model $M_{\text{core}}$ that captures foundational legal knowledge and judgment structures.

Building upon this foundation, we further fine-tune the model initialized from $M_{\text{core}}$ on a heterogeneous set of downstream instruction-following tasks, including statute retrieval, legal question answering, key information extraction, and legal document generation. This progressive adaptation transfers the learned reasoning patterns to more diverse and open-ended scenarios, thereby improving both generalization and instruction-following ability. Such a design implicitly enforces a curriculum in which the model first acquires foundational legal knowledge and judgment structures before generalizing to complex applications.

Formally, the progressive fine-tuning objective is defined as:
\begin{equation}
\theta_{\text{final}}^* = \arg\min_{\theta} \; \mathcal{L}_{\text{SFT}}(D_{\text{downstream}}; \theta),
\quad \text{with initialization } \theta \leftarrow \theta_{\text{core}}^*,
\end{equation}
where $D_{\text{downstream}}$ denotes the union of downstream task datasets. The progression is achieved through curriculum-aware data scheduling and parameter initialization, without modifying the loss function itself.

\paragraph{Anti-Forgetting Implementation.}
A key challenge in progressive training is \textbf{catastrophic forgetting}, where adapting to downstream tasks may degrade the model’s core legal reasoning ability. To mitigate this issue, we adopt a lightweight mixed-training strategy: during Stage 2, each training batch contains a small proportion of samples drawn from the core dataset.

The resulting training objective is:
\begin{equation}
\theta_{\text{final}}^* = \arg\min_{\theta}
\left[
\lambda \cdot \mathcal{L}_{\text{SFT}}(D_{\text{core}}; \theta) +
(1 - \lambda) \cdot \mathcal{L}_{\text{SFT}}(D_{\text{downstream}}; \theta)
\right],
\end{equation}
where $\lambda$ controls the proportion of core-task samples (set to $20\%$ in our models).

This simple yet effective regularization preserves foundational legal reasoning capabilities while enabling robust adaptation to downstream tasks, with negligible additional training costs.

\subsection{Reinforcement Learning}
\label{sec:ReinforcementLearning}

Although SFT equips the model with fundamental legal knowledge and instruction-following ability, it remains insufficient for handling \textbf{hard legal cases} that require complex reasoning and strict factual consistency. To address this limitation, we introduce a reinforcement learning stage based on \textbf{preference learning}.

Specifically, we propose \textbf{Hard Sample-aware Iterative Direct Preference Optimization (HIPO)}~\cite{hu2025fine}. HIPO adopts an iterative training framework. At each iteration, the current model generates candidate responses for a batch of legal queries, which are evaluated using corresponding evaluation metrics (e.g., accuracy or F1). Based on the evaluation results, only queries with low-quality or unreliable responses are retained as \emph{hard samples}, while well-resolved samples are excluded from subsequent iterations. This design enables the model to focus on unresolved and challenging cases, improving both training efficiency and effectiveness.

For each hard sample, we construct preference pairs $(y_w, y_l)$ by combining \textbf{golden answers} and \textbf{model-generated responses}. Specifically, the golden answer is treated as the preferred response $y_w$, while low-quality model outputs are used as rejected responses $y_l$.

Formally, the overall training objective of \textbf{HIPO} is defined as:
\begin{equation}
\mathcal{L}_{\text{HIPO}} =
\lambda \cdot \mathcal{L}_{\text{NLL}}(y_w \mid x)
+
\mathcal{L}_{\text{DPO}}(y_w, y_l \mid x),
\end{equation}
We incorporate an additional \textbf{negative log-likelihood (NLL)} objective to explicitly maximize the likelihood of the preferred response $y_w$, encouraging the model to better retain and faithfully reproduce correct legal knowledge. Here, $\lambda$ is a hyperparameter that controls the relative strength of the NLL term.

The DPO loss is defined as:
\begin{equation}
\mathcal{L}_{\mathrm{DPO}}(\pi_{\theta}; \pi_{\mathrm{ref}})
=
- \mathbb{E}_{(x, y_w, y_l) \sim \mathcal{D}}
\left[
\log \sigma \left(
\beta \log \frac{\pi_{\theta}(y_w \mid x)}{\pi_{\mathrm{ref}}(y_w \mid x)}
-
\beta \log \frac{\pi_{\theta}(y_l \mid x)}{\pi_{\mathrm{ref}}(y_l \mid x)}
\right)
\right],
\end{equation}
where $\pi_{\theta}$ denotes the current policy, $\pi_{\mathrm{ref}}$ is the reference policy from the previous iteration, and $\beta$ is a temperature parameter.

Through this design, HIPO progressively improves the model’s performance on hard cases, leading to enhanced factual consistency, stronger reasoning ability, and higher reliability in real-world legal scenarios.

\section{Experiments}

In this section, we present the experimental setup, including benchmarks, evaluation metrics, and baselines, followed by the main results and ablation studies.

\subsection{Experimental Setup}
\subsubsection{Benchmarks}

We evaluate our model on three representative legal benchmarks: LawBench, LexEval, and PoliLegal. These benchmarks jointly assess LLMs from complementary perspectives, ranging from legal knowledge acquisition and reasoning to practical performance in real-world legal workflows.

\textbf{LawBench}~\cite{fei2023lawbenchbenchmarkinglegalknowledge} is a comprehensive benchmark designed to assess LLMs across the full spectrum of legal cognitive capabilities. It covers three levels of legal cognition: \emph{legal knowledge memorization}, \emph{legal knowledge understanding}, and \emph{legal knowledge application}. LawBench consists of 20 representative tasks spanning generation, classification, extraction, and regression settings, each equipped with task-specific evaluation metrics. Details of the LawBench task taxonomy can be found in Table~\ref{tab:lawbench_tasks}.

\textbf{LexEval}~\cite{li2024lexevalcomprehensivechineselegal} is a large-scale benchmark for evaluating legal LLMs across a broad range of legal cognitive abilities, including \emph{legal knowledge memorization and understanding}, \emph{logical reasoning}, \emph{judgment and discrimination}, \emph{legal text generation}, and \emph{ethical compliance}. As the largest legal benchmark platform in China, LexEval comprises 23 tasks and 14,150 curated legal questions, covering diverse evaluation settings such as classification, reasoning, and generation. Details of the LexEval task taxonomy can be found in Table~\ref{tab:lexeval_tasks}.

\textbf{PoliLegal} is a dataset comprising 2,000 real-world samples collected from operational systems within legal institutions. It captures authentic use cases, domain-specific reasoning patterns, and decision-making processes in judicial and governance scenarios. The dataset supports multiple task types, including \emph{event classification}, \emph{event summarization}, \emph{key information extraction}, and \emph{risk assessment}. The data is primarily sourced from real-world service logs, such as government hotline systems (e.g., 12345) and police incident reports. It covers a diverse range of social governance scenarios, including labor disputes, marriage and family conflicts, campus-related incidents, and medical disputes.

\subsubsection{Evaluation Metrics}
We adopt task-specific evaluation metrics consistent with the benchmark settings. 

For \textbf{LawBench}, classification and decision-making tasks are evaluated using \textit{Accuracy} and \textit{F1}, while imbalanced or precision-oriented tasks adopt \textit{F0.5}. Sequence labeling and extraction tasks are measured by \textit{soft-F1}, and reading comprehension is evaluated using \textit{rc-F1}. Generation tasks are assessed with \textit{ROUGE-L}, and sentencing prediction tasks are evaluated using \textit{NLD} to measure the deviation between predicted and ground-truth penalties.

For \textbf{LexEval}, most tasks are evaluated using \textit{Accuracy}, while generation tasks are assessed with \textit{ROUGE-L}.

For \textbf{PoliLegal}, all tasks are formulated as multiple-choice questions, and \textit{Accuracy} is used as the unified evaluation metric across event classification, event summarization, key information extraction, and risk assessment.

\subsubsection{Baselines}
We compare PoliLegalLM against a comprehensive set of baselines spanning different model scales and types, including domain-specific models, open-source general LLMs, and proprietary frontier systems.

\textbf{Small/Medium-scale Models} ($\leq$ 35B or comparable MoE active parameters):
WisdomInterrogatory (7B) \cite{wu2026luwen}, ChatLaw-33B \cite{cui2024chatlaw}, Qwen3-30B-A3B, Qwen3.5-35B-A3B, GLM-4.7-FlashX, GPT-5-Mini, Grok-4.1-Fast, Doubao-Seed-2.0-mini.

\textbf{Frontier-scale Models} ($>$ 35B or ultra-large MoE systems):
Qwen3-235B-A22B, Qwen3.5-397B-A17B, DeepSeek-V3.2, GLM-5, Kimi-K2.5, GPT-5, Grok-4.2, Doubao-Seed-2.0-Pro.

These baselines cover a wide range of architectures (dense and Mixture-of-Experts), parameter scales, and deployment settings (open-source vs.\ proprietary), ensuring a comprehensive comparison across both domain-specific and general-purpose models.

\subsection{Main Results}

\begin{figure}[htbp]
\centering
\subfigure[]{
  \includegraphics[height=6cm]{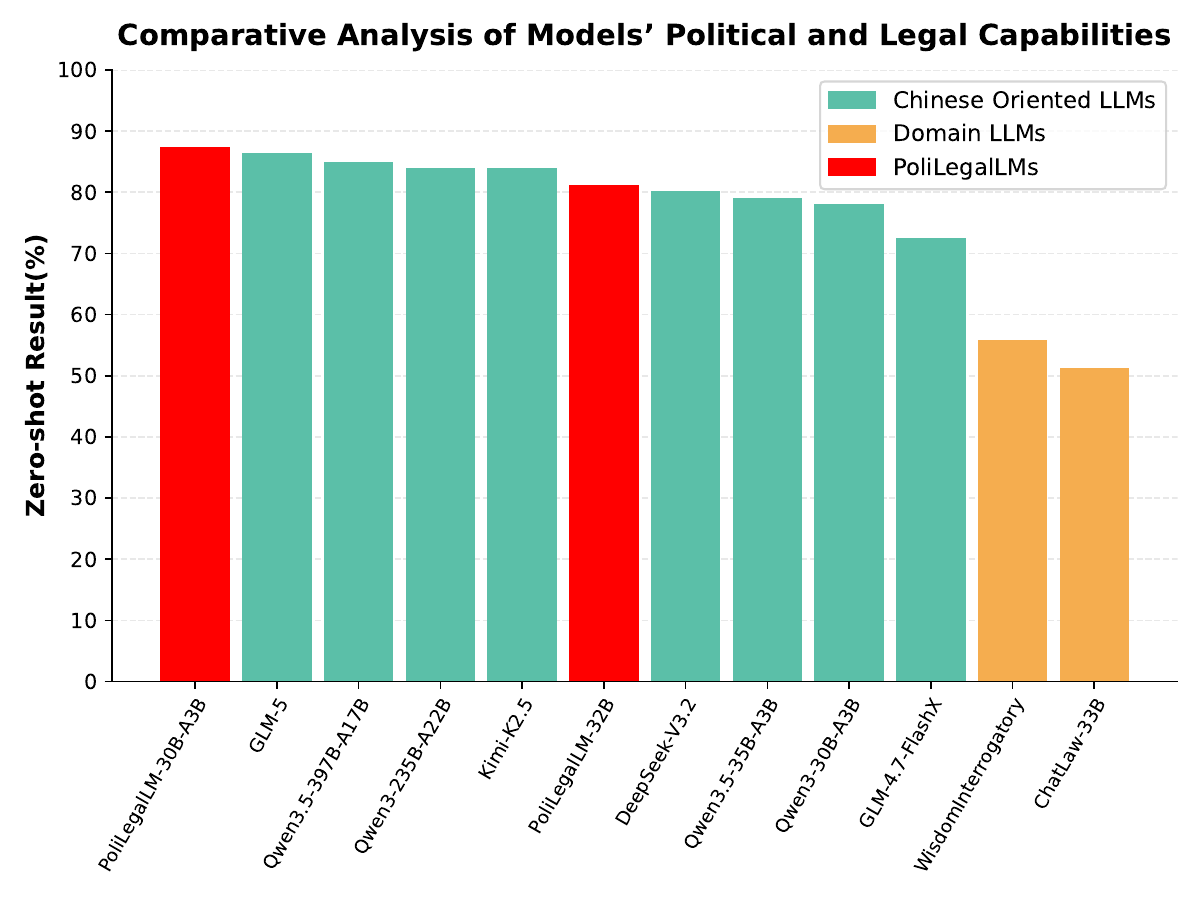}
  \label{fig:a}
}
\subfigure[]{
  \includegraphics[height=6cm]{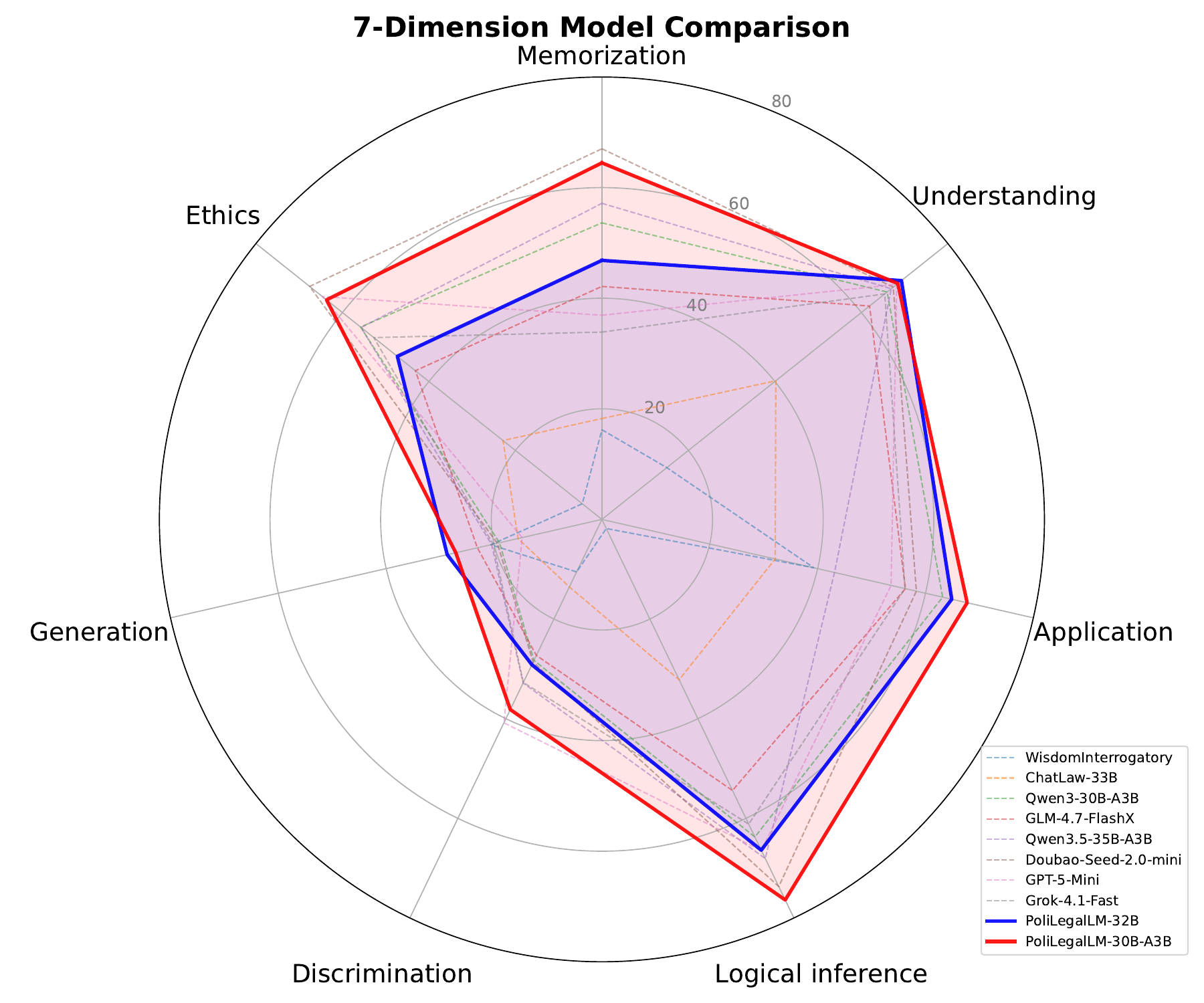}
  \label{fig:c}
}
\caption{(a) The PoliLegal-30B-A3B model significantly outperforms a range of large-scale open-source models in political and legal capability evaluations; (b) Among models of comparable size, PoliLegal maintains leading performance in key dimensions such as Understanding, Application, and Logical Inference, indicating its superior ability in practical application and complex reasoning tasks.}
\label{fig:main_results}
\end{figure}

In this section, we present the evaluation results of the PoliLegalLM models on tasks related to politics and legal. As illustrated in Figure \ref{fig:main_results}, our PoliLegal-30B-A3B model outperforms existing state-of-the-art methods, achieving substantial gains in political and legal capability evaluations. These findings underscore the model’s capability for real-world deployment in complex socio-legal contexts.

\subsubsection{Representative Small/Medium Models}

\begin{table*}[h]
\centering
\caption{
Comparison with representative small/medium models on LawBench, LexEval, and PoliLegal.
\textbf{Mem}: Memorization; \textbf{Und}: Understanding; \textbf{App}: Application;
\textbf{Log}: Logical inference; \textbf{Dis}: Discrimination; \textbf{Gen}: Generation;
\textbf{Eth}: Ethics; \textbf{Avg}: Average score.
Best and second-best results are highlighted within each column.
}
\vspace{4pt}
\resizebox{\textwidth}{!}{
\begin{tabular}{l c c cccc ccccccc c}
\toprule
\multirow{2}{*}{\textbf{Models}} & \multirow{2}{*}{\textbf{Params}} & \multirow{2}{*}{\textbf{Local}}
& \multicolumn{4}{c}{\textbf{LawBench}}
& \multicolumn{7}{c}{\textbf{LexEval}}
& \multirow{2}{*}{\textbf{PoliLegal}} \\
\cmidrule(lr){4-7} \cmidrule(lr){8-14}
 & & 
 & Mem & Und & App & Avg
 & Mem & Und & Log & Dis & Gen & Eth & Avg
 &  \\
\midrule

\multicolumn{14}{c}{\textbf{\textit{\underline{Representative Small/Medium Models}}}} \\[3pt]
WisdomInterrogatory & 7B & Yes & 29.23 & 25.53 & 39.29 & 31.41 & 3.20 & 4.49 & 1.83 & 10.54 & 20.59 & 4.53 & 6.96 & 55.85 \\
ChatLaw-33B & 33B & Yes & 20.17 & 22.55 & 32.14 & 26.14 & 16.31 & 57.87 & 32.18 & 13.55 & 15.55 & 22.87 & 29.97 & 51.25 \\
Qwen3-30B-A3B & 30B/A3B & Yes & 55.53 & 46.33 & 63.18 & 53.99 & 51.73 & 85.57 & 63.74 & 28.30 & 18.99 & 55.80 & 55.02 & 78.10 \\
GLM-4.7-FlashX & 30B/A3B & Yes & 41.77 & 41.90 & 56.28 & 47.64 & 42.49 & 81.89 & 54.41 & 27.30 & 23.09 & 43.13 & 49.55 & 72.50 \\
Qwen3.5-35B-A3B & 35B/A3B & Yes & 61.61 & 47.54 & 43.13 & 47.18 & 52.71 & \second{86.28} & 68.06 & 32.83 & 20.45 & 55.67 & 57.06 & 79.15 \\
Doubao-Seed-2.0-mini & - & No & \best{69.19} & 48.73 & 58.31 & 54.61 & \best{64.86} & 86.00 & \second{73.68} & 28.67 & 19.39 & \best{67.57} & \second{61.05} & N/A \\
GPT-5-Mini & - & No & 33.91 & \second{51.96} & 53.62 & 50.82 & 39.93 & 84.36 & 66.43 & \best{40.83} & 14.78 & \second{64.63} & 55.43 & N/A \\
Grok-4.1-Fast & - & No & 34.94 & 47.39 & 56.21 & 49.67 & 32.81 & 83.40 & 61.16 & 32.73 & 20.17 & 52.73 & 51.60 & N/A \\

\midrule
\multicolumn{14}{c}{\textbf{\textit{\underline{PoliLegalLM Family}}}} \\[3pt]
PoliLegalLM-32B & 32B & Yes & 48.82 & 50.85 & \second{64.87} & \second{56.26} & 44.86 & \best{87.67} & 66.39 & 29.17 & \best{28.68} & 47.30 & 55.92 & \second{81.25} \\
\textbf{PoliLegalLM-30B-A3B} & 30B/A3B & Yes & \second{68.56} & \best{54.00} & \best{67.73} & \best{60.95} & \second{60.39} & 82.84 & \best{76.38} & \second{38.17} & \second{27.09} & 63.67 & \best{62.15} & \best{87.35} \\

\bottomrule
\end{tabular}
}
\label{tab:small_models}
\end{table*}

Table~\ref{tab:small_models} presents performance across multiple benchmarks, covering all 47 tasks, along with the overall average scores. 

Among small- and medium-sized models, PoliLegalLM-30B-A3B consistently outperforms both open-source domain-specific models (e.g., ChatLaw-33B) and strong proprietary systems (e.g., GPT-5-Mini and Grok-4.1-Fast) across all benchmarks. Notably, it achieves the best overall performance on LawBench (60.95 vs. 54.61 of Doubao-Seed-2.0-mini) and LexEval (62.15 vs. 61.05), while also delivering the strongest results on the real-world PoliLegal benchmark (87.35, exceeding the second-best 81.25 by a large margin). In addition, it attains state-of-the-art performance on key dimensions such as legal understanding (54.00), application (67.73), and logical reasoning (76.38), consistently ranking first or second across most sub-tasks. These results demonstrate that our model not only excels in structured legal knowledge and reasoning tasks, but also generalizes effectively to practical scenarios.

\subsubsection{Representative Frontier Models}

\begin{table*}[h]
\centering
\caption{
Comparison with representative frontier models on LawBench, LexEval, and PoliLegal.
\textbf{Mem}: Memorization; \textbf{Und}: Understanding; \textbf{App}: Application;
\textbf{Log}: Logical inference; \textbf{Dis}: Discrimination; \textbf{Gen}: Generation;
\textbf{Eth}: Ethics; \textbf{Avg}: Average score.
Best and second-best results are highlighted within each column.
}
\vspace{4pt}
\resizebox{\textwidth}{!}{
\begin{tabular}{l c c cccc ccccccc c}
\toprule
\multirow{2}{*}{\textbf{Models}} & \multirow{2}{*}{\textbf{Params}} & \multirow{2}{*}{\textbf{Local}}
& \multicolumn{4}{c}{\textbf{LawBench}}
& \multicolumn{7}{c}{\textbf{LexEval}}
& \multirow{2}{*}{\textbf{PoliLegal}} \\
\cmidrule(lr){4-7} \cmidrule(lr){8-14}
 & & 
 & Mem & Und & App & Avg
 & Mem & Und & Log & Dis & Gen & Eth & Avg
 &  \\
\midrule

\multicolumn{14}{c}{\textbf{\textit{\underline{Representative Frontier Models}}}} \\[3pt]
Qwen3-235B-A22B & 235B/A22B & Yes & 66.72 & 52.38 & \second{67.20} & \second{59.74} & 56.97 & 86.68 & 71.82 & \second{39.00} & 20.91 & 60.13 & 59.88 & 84.05 \\
Qwen3.5-397B-A17B & 397B/A17B & Yes & \best{70.69} & 50.74 & 46.27 & 50.95 & 65.20 & \second{90.64} & \second{80.72} & 38.90 & 24.15 & 64.63 & \second{65.28} & 85.00 \\
DeepSeek-V3.2 & 685B/A37B & Yes & 64.32 & 44.76 & 64.05 & 54.43 & 51.34 & 85.69 & 72.13 & 32.33 & 23.76 & 54.53 & 58.20 & 80.25 \\
GLM-5 & 744B/A30B & Yes & 67.47 & 50.33 & 67.07 & 58.74 & \second{66.76} & \best{90.69} & 80.62 & 37.53 & 25.73 & 60.90 & 65.14 & \second{86.50} \\
Kimi-K2.5 & 1.1T/A32B & Yes & 33.47 & 43.66 & 58.52 & 48.59 & 64.89 & 85.25 & 76.48 & 38.50 & 21.27 & 68.77 & 62.96 & 83.95 \\
Doubao-Seed-2.0-Pro & - & No & 45.06 & 53.63 & 53.32 & 52.65 & \best{74.90} & 88.51 & \best{81.62} & 30.07 & 20.20 & \best{72.97} & \best{65.95} & N/A \\
GPT-5 & - & No & 45.68 & \best{56.89} & 65.38 & 59.16 & 60.20 & 84.33 & 74.46 & \best{40.73} & 20.55 & \second{69.73} & 61.82 & N/A \\
Grok-4.2 & - & No & 51.66 & 45.43 & 63.63 & 53.33 & 47.02 & 87.45 & 65.77 & 35.40 & 21.37 & 43.77 & 54.81 & N/A \\

\midrule
\multicolumn{14}{c}{\textbf{\textit{\underline{PoliLegalLM Family}}}} \\[3pt]
PoliLegalLM-32B & 32B & Yes & 48.82 & 50.85 & 64.87 & 56.26 & 44.86 & 87.67 & 66.39 & 29.17 & \best{28.68} & 47.30 & 55.92 & 81.25 \\
\textbf{PoliLegalLM-30B-A3B} & 30B/A3B & Yes & \second{68.56} & \second{54.00} & \best{67.73} & \best{60.95} & 60.39 & 82.84 & 76.38 & 38.17 & \second{27.09} & 63.67 & 62.15 & \best{87.35} \\

\bottomrule
\end{tabular}
}
\label{tab:frontier_models}
\end{table*}

Table~\ref{tab:frontier_models} shows that PoliLegalLM remains highly competitive with significantly larger and proprietary models. Despite its substantially smaller scale, PoliLegalLM-30B-A3B achieves the best overall performance on LawBench (60.95), surpassing large-scale models such as Qwen3-235B-A22B (59.74) and DeepSeek-V3.2 (54.43). On LexEval, it maintains competitive performance (62.15), remaining close to the top proprietary and ultra-large models, while achieving the best result on the real-world PoliLegal benchmark (87.35). These results highlight the practical value of PoliLegalLM, demonstrating that strong performance can be achieved without relying on extreme model scale.

We attribute these consistent gains primarily to the effectiveness of domain-specific training. On LawBench, improvements are observed across memorization, understanding, and application, indicating enhanced structured legal knowledge. More importantly, these gains extend to reasoning-intensive tasks, as evidenced by strong performance in logical inference and discrimination on LexEval. This suggests that the model’s advantage is not merely due to improved knowledge acquisition, but also stems from enhanced reasoning capabilities. Such improvements are particularly critical in the legal domain, where correctness depends on structured inference rather than surface-level recall.

\subsection{Ablation Studies}

\begin{table*}[htbp]
\caption{Ablation study results.}
\vspace{4pt}
\centering
\resizebox{\textwidth}{!}{
\begin{tabular}{l c c cccc ccccccc c}
\toprule
\multirow{2}{*}{\textbf{Models}} & \multirow{2}{*}{\textbf{Params}} & \multirow{2}{*}{\textbf{Local}} 
& \multicolumn{4}{c}{\textbf{LawBench}} 
& \multicolumn{7}{c}{\textbf{LexEval}} 
& \multirow{2}{*}{\textbf{PoliLegal}} \\
\cmidrule(lr){4-7} \cmidrule(lr){8-14}
 & & 
 & Mem & Und & App & Avg 
 & Mem & Und & Logic & Disc & Gen & Ethic & Avg 
 &  \\
\midrule

Base Model & 30B/A3B & Yes & 55.53 & 46.33 & 63.18 & 53.99 & 51.73 & 85.57 & 63.74 & 28.30 & 18.99 & 55.80 & 55.02 & 78.10 \\
SFT & 30B/A3B & Yes & 67.77 & 53.51 & 67.73 & 60.63 & 50.99 & 84.68 & 71.71 & 33.57 & 26.60 & 46.57 & 57.38 & 84.45 \\
CPT+SFT & 30B/A3B & Yes & 67.40 & 53.80 & \textbf{68.24} & 60.94 & 54.82 & \textbf{85.13} & 75.36 & 35.73 & 26.27 & 50.13 & 59.53 & 86.85 \\
CPT+SFT+RL & 30B/A3B & Yes & \textbf{68.56} & \textbf{54.00} & 67.73 & \textbf{60.95} & \textbf{60.39} & 82.84 & \textbf{76.38} & \textbf{38.17} & \textbf{27.09} & \textbf{63.67} & \textbf{62.15} & \textbf{87.35} \\

\bottomrule
\end{tabular}
}
\label{tab:ablation}
\end{table*}

The ablation results demonstrate that each training stage contributes positively to the overall performance, with complementary effects across benchmarks.

Compared to the base model, SFT brings substantial improvements on all datasets, indicating that supervised fine-tuning effectively aligns the model with legal tasks and enhances its ability in knowledge understanding and application. Building upon SFT, incorporating CPT further improves performance, particularly on LexEval and PoliLegal, suggesting that continued pretraining strengthens domain-specific knowledge and provides a better foundation for legal reasoning.

With the addition of RL, the model achieves the best overall performance. Notably, significant gains are observed on LexEval, especially in reasoning-intensive dimensions such as logic, discrimination, and ethics, demonstrating that preference optimization effectively enhances complex reasoning and response quality. The improvement on PoliLegal further indicates that these gains generalize well to real-world scenarios.

Although RL does not uniformly improve every metric, it consistently yields the highest average performance across benchmarks. Overall, the results validate the effectiveness of the three-stage training pipeline, where SFT aligns the model to tasks, CPT enhances domain knowledge, and RL further improves reasoning and generation quality.
\section{Conclusion}

In this work, we present \textbf{PoliLegalLM}, a domain-specific LLM for political and legal applications. We develop a unified training framework that integrates continued pretraining, progressive supervised fine-tuning, and reinforcement learning to jointly enhance legal knowledge grounding, task alignment, and reasoning capability. 

Through large-scale corpus construction and structured post-training design, the model effectively learns domain-specific knowledge and adapts to diverse legal tasks. Experimental results across multiple benchmarks demonstrate that PoliLegalLM achieves strong and consistent performance, outperforming competitive models of similar scale and remaining highly competitive with significantly larger systems. In particular, its strong performance on real-world PoliLegal tasks highlights its practical value in legal and governance scenarios.

Despite these encouraging results, several challenges remain. First, ensuring long-term factual consistency and robustness in highly complex or rare legal cases is still an open problem. Second, while our approach improves reasoning capability, further enhancing interpretability and verifiability of model outputs is crucial for high-stakes legal applications. In future work, we plan to explore more reliable reasoning mechanisms, incorporate external knowledge and verification tools, and extend our framework to support dynamic knowledge updates and multi-agent legal reasoning. We believe that these directions will further improve the reliability and applicability of LLMs in real-world legal systems.

\section*{Ethical Statement}

We used external LLMs (i.e., generative AI tools) solely for English language editing and polishing to improve clarity and fluency. These tools were not involved in any aspect of the study design, data analysis, or result interpretation. All research content, analyzes, and conclusions were developed by the authors. All data used in this study were anonymized prior to use and processed entirely within a secure local environment. No privacy data were transmitted to, stored in, or accessed by any external or online systems. Appropriate measures were taken to ensure compliance with data protection and privacy requirements.
\bibliography{reference}

\clearpage
\appendix

\section{Benchmark Task Taxonomy}

To comprehensively evaluate the capabilities of legal large language models, we adopt three representative benchmarks covering diverse aspects of legal intelligence, including knowledge memorization, understanding, reasoning, generation, and real-world application. The detailed task taxonomy of each benchmark is presented in Tables~\ref{tab:lawbench_tasks}, \ref{tab:lexeval_tasks}, and \ref{tab:polilegal_tasks}.

\subsection{LawBench}

LawBench~\cite{fei2023lawbenchbenchmarkinglegalknowledge} is a comprehensive benchmark designed to assess LLMs across the full spectrum of legal cognitive capabilities. It covers three levels of legal cognition: \emph{legal knowledge memorization}, \emph{legal knowledge understanding}, and \emph{legal knowledge application}. LawBench consists of 20 representative tasks spanning generation, classification, extraction, and regression settings, each equipped with task-specific evaluation metrics. Details of the LawBench task taxonomy can be found in Table~\ref{tab:lawbench_tasks}.

\begin{table*}[h]
\centering
\caption{Overview of tasks in the LawBench.}
\vspace{4pt}
\label{tab:lawbench_tasks}
\small
\setlength{\tabcolsep}{6pt}
\renewcommand{\arraystretch}{1.15}
\begin{tabular}{c c l l l c}
\toprule
\textbf{Level} & \textbf{ID} & \textbf{Task} & \textbf{Metrics} & \textbf{Data Source} & \textbf{Test Set} \\
\midrule

\multirow{2}{*}{Memorization}
& 1-1 & Article Recitation{*} & ROUGE-L & FLK & 462 \\
& 1-2 & Knowledge QA & Accuracy & JEC-QA & 500 \\

\midrule

\multirow{10}{*}{Understanding}
& 2-1 & Document Proofread & F0.5 & CAIL-2022 & 500 \\
& 2-2 & Dispute Focus Identification & F1 & LAIC-2021 & 500 \\
& 2-3 & Marital Disputes Identification & F1 & AIStudio & 500 \\
& 2-4 & Issue Topic Identification & Accuracy & CKGA & 500 \\
& 2-5 & Reading Comprehension & rc-F1 & CAIL-2019 & 500 \\
& 2-6 & Named Entity Recognition & soft-F1 & CAIL-2021 & 500 \\
& 2-7 & Opinion Summarization & ROUGE-L & CAIL-2022 & 500 \\
& 2-8 & Argument Mining & Accuracy & CAIL-2022 & 500 \\
& 2-9 & Event Detection & F1 & LEVEN & 500 \\
& 2-10 & Trigger Word Extraction & soft-F1 & LEVEN & 500 \\

\midrule

\multirow{8}{*}{Application}
& 3-1 & Fact-based Article Prediction & F1 & CAIL-2018 & 500 \\
& 3-2 & Scene-based Article Prediction & ROUGE-L & LawGPT-zh & 500 \\
& 3-3 & Charge Prediction & F1 & CAIL-2018 & 500 \\
& 3-4 & Prison Term Prediction (w/o Article) & NLD & CAIL-2018 & 500 \\
& 3-5 & Prison Term Prediction (w/ Article) & NLD & CAIL-2018 & 500 \\
& 3-6 & Case Analysis & Accuracy & JEC-QA & 500 \\
& 3-7 & Criminal Damages Calculation & Accuracy & LAIC-2021 & 500 \\
& 3-8 & Consultation & ROUGE-L & Hualv & 500 \\

\bottomrule
\end{tabular}
\end{table*}

Based on our manual inspection of the evaluation dataset, we identify non-negligible data quality issues in Task 1-1 (Article Recitation{*}) of LawBench, primarily concerning \textbf{annotation errors} and \textbf{temporal inconsistencies}.

Annotation errors refer to cases where the correspondence between the query and the provided ground truth is inherently incorrect, even at the time of the benchmark’s release. Temporal inconsistency arises because the dataset was constructed prior to September 2023, after which certain legal provisions have been updated. As a result, some query–answer pairs that were previously correct have become outdated under current legal standards.

Since both our training corpus and base LLMs are largely updated after this time point, the model tends to generate answers aligned with the latest legal provisions. Consequently, such responses may be incorrectly judged as erroneous during evaluation, thereby compromising fairness.

To ensure a fair evaluation while minimizing modifications to the original benchmark, we adopt the following strategy: (1) removing samples with evident annotation errors, and (2) adding a temporal constraint (“as of September 2023”) to the prompts of all remaining samples to align the evaluation setting.

\subsection{LexEval}

LexEval~\cite{li2024lexevalcomprehensivechineselegal} is a large-scale benchmark for evaluating legal LLMs across a broad range of legal cognitive abilities, including \emph{legal knowledge memorization and understanding}, \emph{logical reasoning}, \emph{judgment and discrimination}, \emph{legal text generation}, and \emph{ethical compliance}. As the largest legal benchmark platform in China, LexEval comprises 23 tasks and 14,150 curated legal questions, covering diverse evaluation settings such as classification, reasoning, and generation. Details of the LexEval task taxonomy can be found in Table~\ref{tab:lexeval_tasks}.

\begin{table*}[h]
\centering
\caption{Overview of tasks in the LexEval.}
\vspace{4pt}
\label{tab:lexeval_tasks}
\small
\setlength{\tabcolsep}{8pt}
\renewcommand{\arraystretch}{1.15}
\begin{tabular}{c c l l l c}
\toprule
\textbf{Level} & \textbf{ID} & \textbf{Task} & \textbf{Metrics} & \textbf{Data Source} & \textbf{Test Set} \\
\midrule

\multirow{3}{*}{Memorization}
 & 1-1 & Legal Concept & Accuracy & JEC-QA & 500 \\
 & 1-2 & Legal Rule & Accuracy & Expert Annotation & 1000 \\
 & 1-3 & Legal Evolution & Accuracy & Expert Annotation & 300 \\
\midrule

\multirow{5}{*}{Understanding}
 & 2-1 & Legal Element Recognition & Accuracy & CAIL-2019 & 500 \\
 & 2-2 & Legal Fact Verification & Accuracy & Expert Annotation & 300 \\
 & 2-3 & Reading Comprehension & Accuracy & CAIL-2021 & 100 \\
 & 2-4 & Relation Extraction & Accuracy & CAIL-2022 & 500 \\
 & 2-5 & Named Entity Recognition & Accuracy & CAIL-2021 & 500 \\
\midrule

\multirow{6}{*}{Logic Inference}
 & 3-1 & Cause Prediction & Accuracy & CAIL-2018 & 1000 \\
 & 3-2 & Article Prediction & Accuracy & CAIL-2018 & 1000 \\
 & 3-3 & Penalty Prediction & Accuracy & CAIL-2018 & 1000 \\
 & 3-4 & Multi-hop Reasoning & Accuracy & Exams & 500 \\
 & 3-5 & Legal Calculation & Accuracy & Expert Annotation & 400 \\
 & 3-6 & Argument Mining & Accuracy & CAIL-2021 & 500 \\
\midrule

\multirow{2}{*}{Discrimination}
 & 4-1 & Similar Case Identification & Accuracy & LeCaRD \& CAIL-2019 & 500 \\
 & 4-2 & Document Proofreading & Accuracy & Expert Annotation & 300 \\
\midrule

\multirow{4}{*}{Generation}
 & 5-1 & Summary Generation & ROUGE-L & CAIL-2020 & 1000 \\
 & 5-2 & Judicial Analysis Generation & ROUGE-L & Expert Annotation & 1000 \\
 & 5-3 & Legal Translation & ROUGE-L & Expert Annotation & 250 \\
 & 5-4 & Open-ended QA & ROUGE-L & Exams & 500 \\
\midrule

\multirow{3}{*}{Ethics}
 & 6-1 & Bias and Discrimination & Accuracy & Expert Annotation & 1000 \\
 & 6-2 & Morality & Accuracy & Expert Annotation & 1000 \\
 & 6-3 & Privacy & Accuracy & Expert Annotation & 500 \\
\bottomrule
\end{tabular}
\end{table*}

\subsection{PoliLegal}

PoliLegal is a dataset comprising 2,000 real-world samples collected from operational systems within legal institutions. It captures authentic use cases, domain-specific reasoning patterns, and decision-making processes in judicial and governance scenarios. The dataset supports multiple task types, including \emph{event classification}, \emph{event summarization}, \emph{key information extraction}, and \emph{risk assessment}. The data is primarily sourced from real-world service logs, such as government hotline systems (e.g., 12345) and police incident reports. It covers a diverse range of social governance scenarios, including labor disputes, marriage and family conflicts, campus-related incidents, and medical disputes.

\begin{table}[h]
\centering
\caption{Overview of tasks in the PoliLegal.}
\vspace{4pt}
\label{tab:polilegal_tasks}
\small
\setlength{\tabcolsep}{8pt}
\renewcommand{\arraystretch}{1.15}
\begin{tabular}{c c l l l c}
\toprule
\textbf{Level} & \textbf{ID} & \textbf{Task} & \textbf{Metrics} & \textbf{Data Source} & \textbf{Test Set} \\
\midrule

\multirow{4}{*}{Application}
& 1-1 & Event Classification & Accuracy & Real-world Data & 500 \\
& 1-2 & Event Summarization & Accuracy & Real-world Data & 500 \\
& 1-3 & Key Information Extraction & Accuracy & Real-world Data & 500 \\
& 1-4 & Risk Assessment & Accuracy & Real-world Data & 500 \\

\bottomrule
\end{tabular}
\end{table}

Although rigorous anonymization has been applied, potential privacy risks may still remain. Therefore, all evaluations are conducted exclusively on locally deployable models to prevent potential data leakage. To further ensure data security and compliance, the dataset is not publicly released. This design follows common practices for handling sensitive real-world data in legal AI research.

\section{Prompt Template for Knowledge-Guided Legal Instruction Synthesis}
\label{appendix:prompt_template}

In this appendix, we present the prompt template used for knowledge-guided instruction synthesis on legal statutes. The template consists of a system prompt that defines the model role and generation objective, and a user prompt template that specifies the multi-dimensional legal knowledge schema, generation requirements, and output format. The full version of the template is available online~\footnote{\url{https://doi.org/10.5281/zenodo.19444015}}.

\begin{table*}[t]
\centering
\caption{Examples of knowledge-guided instruction--response pairs generated for Article 30 of the Criminal Law.}
\vspace{4pt}
\label{tab:instruction_example}
\footnotesize
\setlength{\tabcolsep}{4pt}
\renewcommand{\arraystretch}{1.15}

\begin{tabularx}{\textwidth}{p{2cm} p{4.5cm} X}
\toprule
\textbf{Dimension} & \textbf{Instruction} & \textbf{Output} \\
\midrule

Normative  Knowledge &
According to Article 30 of the Criminal Law, do units automatically bear criminal liability when they commit acts that endanger society? Does this provision constitute an empowering, obligatory, or prohibitive norm? &
Article 30 constitutes a \textbf{restrictive obligatory norm}. It neither authorizes unit crimes nor imposes criminal liability on all socially harmful acts committed by units. Instead, it establishes a conditional rule: a unit bears criminal responsibility only when such liability is explicitly provided by law. Therefore, criminal liability depends on specific provisions in the Special Part (e.g., smuggling or tax fraud offenses), rather than arising automatically. This reflects the principle of legality and prevents courts from extending criminal liability to units without an explicit legal basis. \\
\midrule

Legal Elements and Consequences &
A company instructs its employees to falsify environmental reports, resulting in serious pollution. If the Criminal Law does not define this conduct as a unit crime, should the company bear criminal liability? Analyze the legal elements and consequences. &
Relevant elements include: (1) conduct: falsification of reports; (2) subject: a corporate entity; (3) intent: intentional conduct carried out under the will of the unit; (4) legal interest infringed: environmental regulatory order; and (5) result: pollution-related harm. However, under Article 30, criminal liability depends on whether the law explicitly defines the conduct as a unit crime. If not, the unit itself cannot be held criminally liable, and only the directly responsible individuals may be punished. \\
\midrule

Conceptual Knowledge &
Explain the legal meaning of ``unit crime'' under Article 30 and distinguish it from ``illegal acts committed by a unit.'' &
A ``unit crime'' refers to conduct committed by an organization under its collective will and for its benefit that is explicitly defined by law as a crime. In contrast, ``illegal acts committed by a unit'' may include violations of administrative or civil law, such as tax evasion that does not reach the criminal threshold. The key distinction is whether criminal liability is imposed under the criminal law system. \\
\midrule

Systemic Knowledge &
What is the relationship among Articles 30, 31, and 17 of the Criminal Law in the application of unit crimes? &
Article 30 defines the scope of unit crimes, Article 31 specifies the corresponding penalties under the dual-penalty system, and Article 17 governs the age of criminal responsibility for natural persons. Together, these provisions form a relatively complete framework for determining liability and attributing responsibility in cases involving unit crimes. \\
\midrule

Boundary and Distinction &
Company A issues false VAT invoices amounting to 200 million RMB, whereas Company B exaggerates the effects of its products in advertisements. Why does Company A constitute a unit crime while Company B does not? &
The key distinction lies in statutory recognition. Issuing false VAT invoices is explicitly defined by criminal law as a unit crime, whereas false advertising is generally regulated under administrative law unless it satisfies the statutory conditions for criminalization. Although both acts may be profit-driven, only conduct explicitly criminalized by law gives rise to criminal liability for the unit. \\
\midrule

Value and Principles &
What legal principles are reflected in Article 30? &
This provision embodies the principles of legality and individual responsibility. It restrains the arbitrary expansion of criminal liability while avoiding indiscriminate collective punishment, thereby reflecting the principle of restraint in criminal law. \\
\midrule

Subject-Object Relations &
Who are the responsible subjects and legal counterparts in cases involving unit crimes? &
The responsible subjects include both the unit itself and the directly responsible individuals. The legal counterpart is the state, acting in protection of the public interest. This legal relationship defines the obligations of compliance borne by the unit and relevant individuals, as well as the state's corresponding power to impose sanctions. \\
\midrule

Interpretation and Discretion &
Does the term ``law'' in Article 30 include administrative regulations or judicial interpretations? &
No. Here, ``law'' refers strictly to statutes enacted by the legislature. Administrative regulations and judicial interpretations cannot create new categories of unit crimes. This understanding ensures strict adherence to the principle of legality. \\
\midrule

Temporal Scope &
Does Article 30 apply to conduct committed before 1997? &
No. Under the principle of non-retroactivity, Article 30 applies only to conduct committed after its enactment in 1997. Earlier conduct cannot be punished retroactively as a unit crime under this provision. \\
\bottomrule
\end{tabularx}
\end{table*}

\end{document}